\documentclass{article}
\usepackage{kr}

\usepackage{times}
\usepackage{soul}
\usepackage{url}
\usepackage[hidelinks]{hyperref}
\usepackage[utf8]{inputenc}
\usepackage[small]{caption}
\usepackage{graphicx}
\usepackage{amsmath}
\usepackage{amsthm}
\usepackage{booktabs}
\usepackage{algorithm}
\usepackage{algpseudocode}
\usepackage{adjustbox}
\usepackage{booktabs}
\usepackage{caption,color}
\usepackage{multirow}

\title{Visual Ground Truth Construction as Faceted Classification}

\author{%
Fausto Giunchiglia\and
Mayukh Bagchi\and
Xiaolei Diao\\
\affiliations
Department of Information Engineering and Computer Science (DISI) \\
University of Trento, Trento, Italy
\emails
\{fausto.giunchiglia, mayukh.bagchi, xiaolei.diao\}@unitn.it
}

\begin{document}

\maketitle

\begin{abstract}

Recent work in Machine Learning and Computer Vision has provided evidence of systematic \emph{design flaws} in the development of major object recognition benchmark datasets. One such example is ImageNet, wherein, for several categories of images, there are incongruences between the objects they represent and the labels used to annotate them. The consequences of this problem are major, in particular considering the large number of machine learning applications, not least those based on Deep Neural Networks, that have been trained on these datasets. In this paper we posit the problem to be the lack of a knowledge representation (KR) methodology providing the foundations for the construction of these ground truth benchmark datasets. Accordingly, we propose a solution articulated in three main steps: (i) deconstructing the object recognition process in four ordered stages grounded in the philosophical theory of teleosemantics; (ii) based on such stratification, proposing a novel four-phased methodology for organizing objects in classification hierarchies according to their visual properties; and (iii) performing such classification according to the faceted classification paradigm. The key novelty of our approach lies in the fact that we construct the classification hierarchies  from visual properties exploiting \emph{visual genus-differentiae}, and not from linguistically grounded properties. The proposed approach is validated by a set of experiments on the ImageNet hierarchy of musical experiments.
\end{abstract}

\section{Introduction}
\label{S1}
In Machine Learning (ML), and Computer Vision (CV) as a particular case, the models developed are informed by training them on 
annotated datasets, which are supposed to be a \textit{ground truth}, i.e., a high quality objective representation of what is the case in the world. In many cases, e.g., with datasets used to train Deep Neural Networks, the size of these datasets can become very large, some examples being Open Images \cite{openimages}, COCO \cite{coco}, YFCC100M \cite{yfcc100m}, YouTube-8M \cite{Youtube-8m}, NTU RGB+D \cite{Ntu-rgb+d}.
Within this line of work, maybe the most relevant dataset is ImageNet \cite{IMAGENET-2009}, its main strengths being the size (counting millions of photos), the quality, as compared to the others (ImageNet was built by populating WordNet \cite{PWN}), and the fact that, in the last years, it has been used to train some of the most successful Neural Networks, see, e.g., AlexNet \cite{alexnet}, VGGnet \cite{vgg}, GoogleNet \cite{googlenet}, ResNet \cite{resnet}, DenseNet \cite{densenet}, becoming a \textit{de-facto} benchmarking standard. However, lately a major concern has grown about the quality of these ground truth datasets and of the implications on the quality and performance of the resulting ML models, see, e.g., \cite{sambasivan2021everyone,raji2021ai,koch2021reduced} and also \cite{cheng2015semantically} which suggests using WordNet for improving the quality of the labels used to search for and annotate images.

Of specific relevance is the work in \cite{ICML-2020}\footnote{See also the extended arXiv version of the paper at: https://arxiv.org/abs/2005.11295} which focuses on the mistakes in ImageNet. Our basic tenet is that these mistakes are \textit{not} and \textit{simply} the result of carelessness on the side of the annotators, being in fact deeply grounded in the way language and perception interact. 
As a matter of fact, an early version of this type of problems was crystallized, already in 2000, as the \textit{Semantic Gap Problem} (SGP) \cite{SGP-2000}, where the SGP was described 
as the ``\emph{lack of coincidence between the information that one can extract from the visual data and the interpretation that the same data have for a user in a given situation}". 
This problem, still unsolved, has been generalized in \cite{SNCS-2021} as the fact that there is a \emph{many-to-many mapping} between the information extracted from the visual data and its possible contextual interpretations.  As from this work, the SGP is a very general phenomenon which manifests itself anytime an object inside an image is linguistically annotated multiple times, possibly even by the same person. The SGP is in fact a consequence of the fact that  linguistic descriptions of images are subjective and context dependent. \cite{2016-FOIS} provides a detailed analysis, grounded on the Teleosemantics theory of meaning \cite{macdonald2006teleosemantics,millikan2000,millikan1989}, of the mechanisms by which the SGP arises.

The goal of this paper is to provide a general Knowledge Representation (KR) methodology for generating high-quality ground truth datasets. The intuition is that KR can provide  modeling guidelines which will drive the organization of the datasets used to train ML models, thus indirectly inducing in them the \textit{semantics awareness} that they are missing, and of which the SGP is a major evidence. \textit{Instead of explicitly codifying the intended semantics via axioms, the idea is to encode them in the process by which ML models are provided with ground knowledge, in the form of training examples.} In this perspective,
 ImageNet is a very good starting point given that it already organizes images according to a lexico-semantic hierarchy. The issue is to align this hierarchy with the \textit{visual semantics} encoded in the objects depicted in images. To this extent we follow the theory introduced in \cite{SNCS-2021} and associate objects, as depicted in (multiple) images, with sets of \textit{visual properties}, e.g., sets of (visual) frames, which describe their appearance. In this way, the annotation process is no longer that of annotating (objects in) images with labels but, rather, of aligning the visual properties of (objects in) images with those lexically described properties which describe the meaning of  ImageNet classes, i.e., their definition, or \textit{gloss}, at it is called in WordNet \cite{PWN}. 
Labels are then (optionally) associated to ImageNet classes when their definition is consolidated from both a language and a vision point of view.

\begin{figure}[t]
\includegraphics[width=8cm,height=3cm]{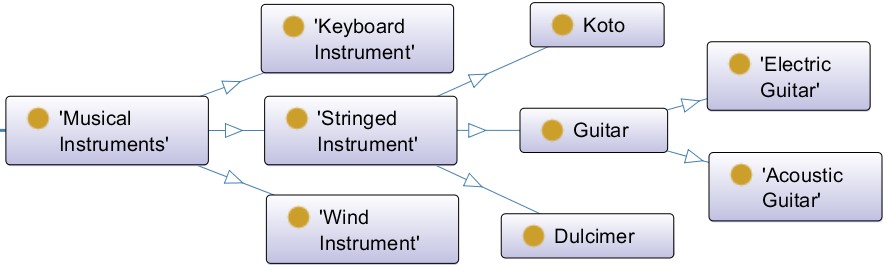}
\caption{An ImageNet Musical Instrument Sub-Hierarchy.}
\centering
\label{I0}
\vspace{-0.5cm}
\end{figure}

We organize the methodology we propose as a stratified four-step annotation process whereby: first (i) the relevant objects in an image are identified, then (ii)  these objects are characterized by their \textit{visual properties}, then (iii) they are linguistically annotated using ImageNet-like labels and, finally, (iv) they are associated a unique identifier. The main contributions of this paper are as follows:
\begin{itemize}
    \item A reconstruction and extension of the annotation mistakes highlighted in \cite{ICML-2020} as specific instances of the SGP.
    \item The definition of a four-step annotation process dealing with the SGP. Of notable relevance is that annotations are done in terms of \textit{Genus-Differentia} of the visual properties of objects, making sure that these visual properties are aligned with the linguistic properties described in glosses. 
    \item The implementation of the four-step annotation process as faceted classification \cite{SRR-67,SRR-89}, suitably adapted to deal with \textit{visual properties} and images \cite{2021-CAOS,2014-DERA}, thus providing precise guidelines (so called \textit{canons} in faceted classification) towards high quality annotation results.
    \item The exploitation of state of the art multi-lingual lexical resources, this allowing, among other things, the possibility of language-aware annotations, well beyond the current practice of English-only annotations \cite{UKC-IJCAI}.
\end{itemize}
\noindent
The paper is organized as follows. In Section \ref{S2} we describe the types of mistakes which arise because of the SGP. In Section \ref{S3} we describe the annotation process and how it deals with the mistakes just mentioned. In Section \ref{S4} we describe how to implement it as faceted classification. In Section \ref{S6} we evaluate the proposed methodology. We do this via an annotation and a machine learning experiment. Section \ref{S7} describes the related work, while Section \ref{S8} concludes the paper.
All along the paper, we use the dataset represented in Fig.\ref{I0}, consisting of nine categories and 3660 images. 

\noindent

\section{The Semantic Gap Problem}
\label{S2}
As from \cite{ICML-2020}, in ImageNet there are three recurring \emph{design flaws} which are associated to three specific types of images. Following this characterization, we categorize the images in ImageNet into three plus one broad categories, where the fourth one is the category of the images that in \cite{ICML-2020} are taken as being properly annotated. We have the following.

\begin{enumerate}
\item \textbf{Good Images}, for which there is wide inter-annotator agreement.

\item \textbf{Multi-Object Images}, where the flaw arises from the occurrence of multiple objects in the same image.

\item \textbf{Single-Object Images}, where the flaw arises due to the assignment of \emph{mutually exclusive} labels to a single object.

\item \textbf{Mislabelled Images}, where the flaw arises due to labelling mistakes.
\end{enumerate}
Evidence that the last three  categories suffer from the SGP is provided by the fact that  the ImageNet creators and the authors of \cite{ICML-2020} have contradictory opinions about them. However, also the Good Images category suffers from the SGP. As a matter of fact, all four categories present the SGP problem as it appears in Good Images, the last three showing an additional form of SGP specific to the class. Let us analyze these four categories in detail.
\begin{figure*}[htp]
\includegraphics[width=18cm,height=4cm]{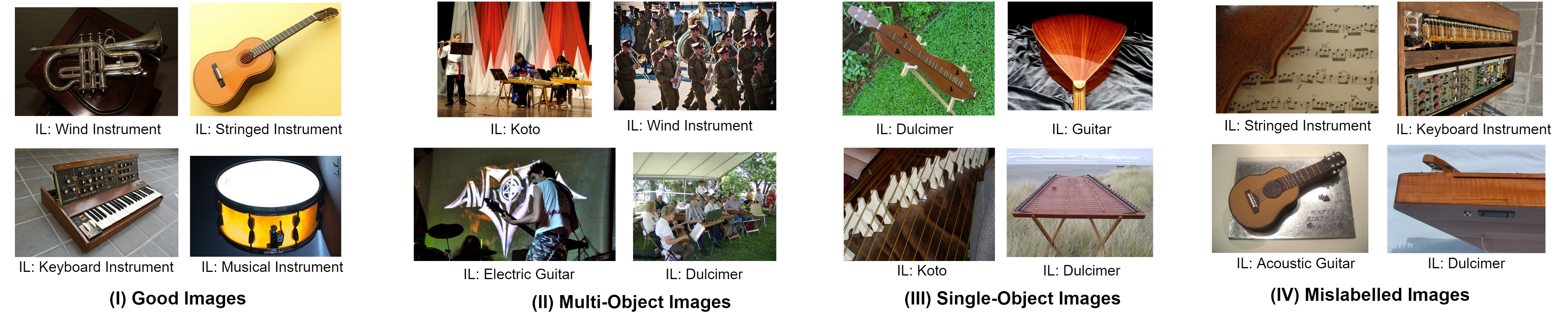}
\vspace{-0.6cm}
\caption{Categorized Image Samples from ImageNet Musical Instrument Sub-Hierarchy}
\vspace{-0.5cm}
\centering
\label{I1}
\end{figure*}

\vspace{0.1cm}
\noindent \textbf{Good Images:} 
We can notice at least three characteristics definitive of what we categorize to be a good image. Firstly, these images are almost always those images containing a single object (the \emph{`main object'} as called in \cite{ICML-2020}). Secondly, these images are less noisy, in the sense that they have minimum influence of confounding variables such as occlusion and clutter distorting them. Thirdly, all of these images are captured from an optimal viewpoint leading to clear visibility of their defining visual characteristic; see, e.g., the image-label pairs for (I) Good Images in Fig.\ref{I1} (`IL' stands for ImageNet Label).

Three observations. Firstly, though it is a fact that confounding variables are ubiquitous in open world settings \cite{bendale2015towards}, the training corpora for object recognition models should be majorly, but not only, comprised of good images. This is crucial given the fact that it is \emph{only} the good images which can allow the model to extract and learn about its distinguishing visual characteristics. Secondly, and again ubiquitous in open world settings, is the issue of \emph{parts} of an object in an image. Studies such as \cite{ICML-2020} have considered different labels for objects and its parts (e.g., \emph{`car'} vs \emph{`car wheel'}) treating them as distinct objects. The present work being focused on methodological visual classification postpones the specific issue of detection of parts of an object to a future version of the methodology. Thirdly,  the SGP still persists for any of these images, even if agreed upon by multiple annotators. To realize this it is sufficient to think of the many other labels we could subjectively use to describe any of the good images in Figure \ref{I1}. (As from above, see \cite{2016-FOIS} for an in depth analysis of this phenomenon.)

\vspace{0.1cm}
\noindent \textbf{Multi-Object Images:} An image in ImageNet is associated with only one label, this causing the first category of design flaw in case of \emph{multi-object images}, i.e. images comprising objects from multiple classes. With such images constituting more than one-fifth of ImageNet's total image populace, the flaw stems from the \emph{systematic incongruence} between the ImageNet label of (each of) such images and the label of the most likely main object in (each of) them as deemed by humans. A famous example is the image of a \emph{`stage'} labelled by humans as such, having the ImageNet (IL) label as \emph{`Electric Guitar'} (see this and more examples in (II) Multi-Object Images of Fig.\ref{I1}). 

Multi-object images assume central importance due to two pivotal observations. Firstly, from empirical evidence in cognitive psychology \cite{ROSCH-1976}, the observation that the main object chosen by humans possesses the highest \emph{`cue validity'}, in other words, carry the most information via perceptual attributes (\emph{stimuli}) and thus is visually the most salient. Secondly, in violation of the first observation, ImageNet exhibits an established bias for many multi-object images wherein, for such an image, its label correspond to a very \emph{distinctive object} instead of the main object in that image, thus exploiting features that don't generalize to object recognition in the wild \cite{ICML-2020}. 

\vspace{0.1cm}
\noindent \textbf{Single-Object Images:} The second category of design flaw concerns \emph{single-object images}. It stems from the empirical observation that humans may assign multiple mutually exclusive labels possibly due to the object in the image being \emph{visually polysemic}\footnote{Identified in \url{https://gradientscience.org/benchmarks/}} or in the case of classes having synonymous labels (such as in \emph{confusing class pairs}). Images are visually polysemic when their \emph{``semantics are described only partially"} \cite{SGP-2000} and their interpretation is not unique. A concrete example (amongst many others) of visual polysemy in ImageNet is the case of an image (see the image of IL: Guitar in (III) Single-Object Images of Fig.\ref{I1}) for which both the labels \emph{`guitar'} and \emph{`bass'} can equally contextually be assigned. \cite{ICML-2020} observes that not only do humans assign an alternative label \emph{``as often as"} the ImageNet label for 40\% of ImageNet images, they also assign as many as up to 10 labels for many single-object images. 

The second case of confusing class pairs occurs when humans are unable to disambiguate pair of (semantically very similar) classes and assign same (set of) labels for either of them. A prominent example in ImageNet are the two images labelled as `IL: Dulcimer' (see (III) Single-Object Images of Fig.\ref{I1}) which, though considered semantically very similar by humans and assigned the ImageNet label \emph{`Dulcimer'} out of ambiguity, essentially belong to two different genres of musical instruments. The study in \cite{ICML-2020} attributes two \emph{design factors} of ImageNet behind such confusing class pairs. Firstly, some of the cases could be due to possible overlap or mixup in their image distributions due to ImageNet's automated image retrieval process. Secondly, and most importantly, choosing disjoint labels grounded in linguistic properties is insufficient for humans to visually disambiguate confusing class pairs in the face of potential overlapping image distributions.

\vspace{0.1cm}
\noindent \textbf{Mislabelled Images:} The third category of design flaw relates to mislabelled images, for (each of) which there is no match between the ImageNet label and the (correct) label(s) assigned by humans. Such images can be grouped into two categories based on their mode of identification. The first category are those for which the selection frequency of the ImageNet label was zero, or, in other words, no human selected (an object corresponding to) the ImageNet label to be \emph{contained} in such images. The second category are those for which the ImageNet label was not even considered by humans for \emph{annotating} any object within such images. A concrete example, for instance, can be seen in Fig.\ref{I1} (See (IV) Mislabelled Images) wherein the image labelled by ImageNet as \emph{`acoustic guitar'} is a `fake' guitar shaped on a birthday cake, and easily identifiable as such. 

\begin{table*}[]
\caption{Image annotation via faceted classification.}
\label{T1}
\centering

\vspace{-0.2cm}
	\resizebox{1\linewidth}{!}{%
\begin{tabular}{|l|l|l|l|l|}
\hline
\textit{\textbf{Annotation Stage}} & \textit{\textbf{Teleosemantics Grounding}} & \textit{\textbf{Faceted Classification}} & \textit{\textbf{Problem}} & \textit{\textbf{Solution}}  \\
\hline
S1: Object Detection          & Generation and Modelling of SCs & Pre-Idea Stage & Multi-Object Images & Bounding Polygons \\
S2: Visual Classification     & Hierarchy Construction of SCs & Idea Plane & All Images SGP  & Visual Genus-Differentia   \\
S3: Linguistic Classification & Linguistically Labelling SCs as CCs & Verbal Plane & Mislabelled Images & Language Labels \\
S4: Conceptual Classification & Alinguistic Rendering of CCs & Notational Plane & Single-Object Images & Alinguistic Identifiers \\
\hline
\end{tabular}}
\vspace{-0.3cm}
\end{table*}

\section{The Four-Stage Annotation Process}
\label{S3}

The proposed stratification of the annotation process is founded on a fundamental distinction, and consequent execution \emph{order} between \textit{object detection}  and \textit{visual (image) classification}. This is a crucial step which is actually systematically \emph{collapsed} and \emph{non-delineated} in state-of-the-art object recognition approaches (see  footnote in page 1 of \cite{ILSVRC}). In this context, by object detection we mean the activity by which all objects in an image are localized (but not identified), for instance via  \emph{bounding polygons} \cite{ILSVRC}. On the other hand, by \emph{image classification}, we mean the activity which determines what object classes are present in an image \cite{ILSVRC}. This latter task, which is \emph{semantics-intensive}, is performed post object detection, wherein, given a continuous feed of images, the goal is to organize the objects in them into a classification hierarchy of \emph{visual objects} based on their \emph{visual genus-differentia} \cite{SNCS-2021}. 



We ground the above ordered distinction in the empirically validated theory of \emph{teleosemantics} \cite{2016-FOIS,2017-teleologies,millikan2020neuroscience}, as originally postulated by the philosopher Ruth Millikan \cite{millikan1989,millikan2000,millikan2004,millikan2005}.
At the outset, we outline two key postulates. Firstly, we model the (part of the) world as being populated by substances which are \emph{``things about which you can learn from one encounter something of what to expect on other encounters, where this is no accident but the result of a real connection”} [Quote from \cite{millikan2000}]. Notice that the notion of substances we commit to here are constrained to only those which can be visually perceived (i.e. \emph{objects}). Secondly, we model concepts generated from substances as \emph{mental abilities} implementing suitable (etiological) \emph{functions} which must be understood as \emph{`intended for’} a specific purpose. Such a modelling paradigm enables us to distinguish between the distinct abilities which drive image classification, e.g., object detection, object recognition and linguistic description. Accordingly, we have \emph{substance concepts} (SCs) focused on (continual) object detection and visual classification from substances, and (ii) \emph{classification concepts} (CCs) geared towards (continual) linguistic classification and description of substance concepts \cite{2019-PRICAI}. 

Based on the above assumptions, we stratify the annotation process (and, therefore, as from above, the recognition process) in four stages, as from Table \ref{T1}. Here, the first column reports the specific annotation stage, while the second maps these stages to the corresponding Teleosemantics activity. Let us consider these stages in  detail.

\vspace{0.1cm}
\noindent \textbf{S1: Object Detection.} Substances are detected over multiple sets of encounters (e.g., sets of images), as \emph{visual objects}, i.e., sets of similar visual frames. Substance concepts are, in turn, modelled as \emph{``sets of visual objects"} used to \emph{``represent substances as they are perceived"}, incrementally, at different levels of abstraction. This process involves (continual) extraction of distinguishing visual properties unique to a substance concept, and thereby delineating substance concepts, for instance, via bounding polygons. Subsequently, substance concepts are stored in a \emph{``cumulative memory M of all the times they were previously perceived"} \cite{SNCS-2021}. 

There are three key features which characterize substance concepts to be perfectly amenable for incremental object detection. Firstly, differently from conventional KR formalisms, the detection process of substances is independent, \emph{per se}, of the distinction between an individual (such as \emph{`guitar\#123'}) or a real kind (such as \emph{`guitar'}). The exact granularity of object detection depends on the purpose of (the user of) the object recognition system. Secondly, the key to continual detection of a substance as different from substance concepts is grounded in its internal \emph{causal factor} \cite{2016-FOIS} which is incrementally manifested and extracted as perceivable visual properties. Finally, and most importantly, substance concepts are \emph{perdurant} representations \cite{2002-DOLCE} of objects as \emph{``we never have a full (visual) picture of the object but that its visual representation is built progressively, in time"} \cite{SNCS-2021}. The perdurant representation is, in fact, grounded in the notions of space and time \emph{persistency} \cite{SNCS-2021} which ensures that even if an object isn't fully visually perceivable at once, it does exhibit very slow spatio-temporal variance which allows for its detection and substance concept representation to be built, progressively and cumulatively, in time.

\vspace{0.1cm}
\noindent \textbf{S2: Visual Classification.} Concurrently with continual object detection, and after new objects are detected, the next step is to build \emph{visual subsumption hierarchies} exploiting  the \emph{visual genus-differentia} as (continually) extracted from the (incrementally) perceivable visual properties of the new images. The key notion of visual subsumption hierarchy, as introduced in \cite{SNCS-2021}, refers to a (dynamic) classification hierarchy of the continually perceived substance concepts learnt, \emph{ab initio}, from visual genus-differentiae. Visual Genus refers to a set of visual properties shared across distinct objects, of which a certain representative object is referred to as the Genus Object. Visual Differentia, on the other hand, refers to a set of novel visual properties different from those of the visual genus, which are exploited to differentiate amongst different objects with the same genus. The key observation is that, for any category, \textit{the differentia one level up, becomes the genus one level down}. In other words, what one level up differentiates a class (of objects) from another, is the common part shared by all objects in the category one level down. This common part then then becomes the new genus, namely the starting point for a new split based on a new and more refined differentia. In other words, differentia is the only key notion that we need when building a visual subsumption hierarchy. For example, given that we take the basic category \emph{`stringed instrument'} as the representative Genus Object with its Visual Genus as the \emph{`presence of taut strings'}, one pre-eminent way of visually classifying further can be based on the Visual Differentia \emph{`the number of taut strings'} with its different instantiations, e.g., \emph{`six taut strings'}, \emph{`thirteen taut strings'} etc. 

The crucial and most important observation is that, during this stage, the many-to-many mapping of the SGP, as it occurs in the Good Images category (see Section \ref{S2}) is caused exactly by the choice of different differentiae for the same genus. And this is also why a classification based on class labels, will generate the SGP many-to-many mapping. It is sufficient that two annotators, or even the same annotator, based on their personal experience, when selecting the class label, implicitly (and without noticing) apply a different differentia to two images of the same object, and the SGP will appear. For instance, the image labelled as \emph{Keyboard Instrument} in Fig.\ref{I1} (I) can be labelled differently as \emph{synthesizer} but also as a \emph{keyboard workstation}, the underlying (though implicit) visual differentia for the former being just the presence of \emph{`keyboard'} (e.g., for a common user) and for the latter the visual presence of the \emph{`control panel'} (e.g., for a musician).


Because of this, in this stage S2, our methodology makes three fundamental assumptions:
\begin{enumerate}
    \item The visual subsumption hierarchy is built based on the visual genus and differentia of objects and \textit{not} on their class labels.
    \item The visual properties on which the differentia is computed are consistent across \textit{all} objects in that category.  
    \item The visual properties used to compute the visual differentia are consistent with the modelling decisions that are taken linguistically, i.e., with the genus and differentia as they are described in the gloss of the corresponding (ImageNet) category.
\end{enumerate}
Notice how our approach is a radical departure from mainstream CV and in particular from how ground truth datasets have been generated so far. Notice also how visual classification, more specifically, the (successive) selection of the visual differentia assumes an \emph{egocentric setting} \cite{2020-ECAI}, in other words, such selection is completely bound to the point-of-view, experience and the purpose of the user. In fact, as evidenced from cognitive psychology \cite{1989-Palmer}, the selection of what we define as visual differentia depends on the highly egocentric differentiation of \emph{affordances} \cite{1977-Gibson} - visual properties which have meaning for function of an object (e.g., the visual property \emph{taut strings} for a musical instrument denotes the function of playing them to produce sound). From this point of view, Wordnet and ImageNet are just the result of a series of subjective choices, the first on the differentia provided linguistically, the latter on the corresponding visual differentia. No claim of universality can be made, we can only strive for achieving the most widely accepted organization of concepts, see also the discussion in \cite{UKC-CICLING}.

\vspace{0.1cm}
\noindent \textbf{S3: Linguistic Classification.} This phase focuses on (continual) linguistic labelling of the represented substance concepts with ground truth labels as soon as the corresponding objects are detected and visually classified. This results in also a continual conversion of substance concepts into classification concepts - the linguistic description of substance concepts - which, differently from substance concepts, acts as vehicles for formal communication and reasoning \cite{2019-PRICAI}. 

Notice, however, that this phase is \emph{non-trivial} with respect to at least four decisive aspects. Firstly, the fact that linguistic phenomena such as \emph{synonymy} and \emph{polysemy} induce a many-to-many mapping between substance concepts and classification concepts, thus, resulting in a combinatorial explosion of ground truth labels to choose from, both within and among different natural languages as well as domain languages. Secondly, the crucial influence of \emph{lexical gaps} \cite{UKC-IJCAI,UKC-CICLING} on (cross-lingual) object recognition, namely the inviolable fact that a substance concept can be recognized \emph{if and only if} it has \emph{at least one} corresponding classification concept (e.g., the label \emph{`koto'} denoting a musical instrument is a lexical gap in Bengali language). Thirdly, as amply exemplified in \cite{2019-CVPR}, the fact that even for a single concept (e.g., \emph{marriage}), the visual stimuli (e.g., images) can be radically diverse depending on the language (and ultimately, culture) via which it is visually conceptualized (e.g., images of marriages as conceptualized in English vs. Hindi \cite{2019-CVPR}). Last but not the least, differently from mainstream CV where visual classification itself is via labels \emph{only}, we keep visual classification (effectuated via visual genus-differentiae) \emph{characteristically distinct} (but, functionally linked) from linguistic ground truth labelling.

\vspace{0.1cm}
\noindent \textbf{S4: Conceptual Classification.} The focus of the last stage of our stratified approach, conceptual classification, is to render linguistically grounded classification concepts \emph{alinguistic}. The ultimate goal is to represent each (word sense of the) ground truth label linked to a unique classification concept (which, in turn, is linked to several images) as a \emph{language independent} or \emph{alinguistic} numerical identifier (as a result of which, linguistic phenomena such as polysemy and synonymy are tackled). 

This stage is crucial due to the following three observations.
Firstly, our solution approach attempts to seamlessly integrate semantically equivalent ground truth labels across multiple languages (inclusive of both natural languages and domain languages). Thus, in a nutshell, it ventures beyond mainstream object recognition systems exhibiting a \emph{unilingual bias}, mostly, towards the English language as detailed in \cite{2019-CVPR}. Secondly, it transcends multiple cultures, in the sense that, for an object recognition system to be representative and fair \cite{2019-CVPR}, it must accommodate conceptual hierarchies composed of different levels of abstraction, primarily, but not only, due to different genres of \emph{representation diversity} \cite{2020-ETR-KR} pervasive across cultures and domains. Finally, as a consequence of the above two observations, it is important to notice that our approach accommodates and links together representations of multiple ground truth hierarchies as experienced via different languages and cultures.

The four stages above  allow to deal with the design flaws described in section \ref{S2}. The fourth column in Table \ref{T1} reports the flaw while the fifth describes the specific solution. Firstly, the crucial highlight that the object detection stage especially tackles the problem of \emph{multi-object images} as it extracts (features of) different substance concepts and uniquely delineates (each of) them via bounding polygons. For instance, in Fig.\ref{I1} (II), via object detection, the objects in the image labelled as \emph{Koto} can be delineated via bounding polygons to be not only comprising of \emph{koto} but also \emph{flute}, \emph{music stand} etc. This ensures the rejection of very poor quality \emph{`confounding'} multi-object images as ground truth. Secondly, the visual classification stage implemented via exploiting visual genus-differentiae of (delineated) objects is equally pivotal for \emph{all image categories SGP} (including good images). It eliminates those images (e.g, the image labelled \emph{electric guitar} in Fig.\ref{I1} (II)) as ground truth where the visual differentia is opaque. The linguistic classification stage, on the other hand, provides a specific solution to the \emph{mislabelled images} because it offers a repertoire of natural language as well domain vocabularies from which to choose appropriate language labels for annotation. For instance, the guitar shaped cake in Fig.\ref{I1} (IV) will never be labelled as an \emph{acoustic guitar} following linguistic classification. The last stage of conceptual classification provides a unified solution for all image categories via the assignment of a unique alinguistic identifier to a visual concept, which \emph{mutadis mutandis}, replicate the same notion in lexical semantics to absorb polysemy. This stage especially tackles the visually polysemic \emph{single-object images} such as the two images labelled as dulcimers in Fig.\ref{I1} (III) as, following conceptual classification, the visual concept of dulcimer will have a unique alinguistic meaning and hence, \emph{at most}, one of the aforementioned images can be a dulcimer.

Finally, even though the four-stage stratified process above provides a teleosemantically well-founded mechanism for modelling media annotations, the SGP many-to-many problem still persists. For example, let us take the \emph{simplest} case of the (good) image \emph{``stringed instrument"} (Fig.\ref{I1} (I)) as the visual data. The process above doesn't provide any guiding principle(s) which can enforce a \emph{high quality} explicit selection, successive application and hierarchical modelling of visual differentiae from visual data. The first consequence of this is the (open) possibility to conceptualize radically diverse substance concepts from the image (such as \emph{acoustic guitar} but also \emph{ochre-colored guitar} etc.), each of which can further be visualized as diverse images. The second consequence is the fact that even for a single substance concept, the labelling can be done differently in diverse languages and cultures, the precise semantics of which might be similar but \emph{not necessarily} the same. E.g., the \emph{acoustic guitar} can be labelled variously as \emph{hawaiian guitar}, \emph{non-electric guitar} etc., each of which can further have diverse imagery. This is exactly the problem dealt via faceted classification, as described in the next section. In this perspective \textit{faceted classification can be seen as the general methodology for generating and/ or evaluating the correcteness and/or modifying or extending ImageNet-like datasets.}

\section{The Faceted Classification Process}
\label{S4}


%


\begin{table*}[!t]
\centering
\caption{Statistics about annotation problems in the ImageNet musical instruments sub-hierarchy.}
\vspace{-0.2cm}
\label{tab:2}
	\resizebox{1\linewidth}{!}{%
\begin{tabular}{|l|cc|cc|cc|cc|cc|cc|cc|cc|cc|}
\hline
\multicolumn{1}{|c|}{\multirow{2}{*}{\textbf{Number}}}  & \multicolumn{2}{c|}{\textbf{Musical   Instrument}} & \multicolumn{2}{c|}{\textbf{Stringed   Instrument}} & \multicolumn{2}{c|}{\textbf{Keyboard   Instrument}} & \multicolumn{2}{c|}{\textbf{Wind   Instrument}} & \multicolumn{2}{c|}{\textbf{Guitar}} & \multicolumn{2}{c|}{\textbf{Dulcimer}} & \multicolumn{2}{c|}{\textbf{Koto}} & \multicolumn{2}{c|}{\textbf{Acoustic   Guitar}} & \multicolumn{2}{c|}{\textbf{Electric   Guitar}} \\ \cline{2-19} 
                                 & \textbf{Original}           & \textbf{UE}          & \textbf{Original}           & \textbf{UE}           & \textbf{Original}           & \textbf{UE}           & \textbf{Original}         & \textbf{UE}         & \textbf{Original}    & \textbf{UE}   & \textbf{Original}     & \textbf{UE}    & \textbf{Original}   & \textbf{UE}  & \textbf{Original}         & \textbf{UE}         & \textbf{Original}         & \textbf{UE}         \\ \hline
\textbf{Good Images}             & 175                         & 17                   & 134                         & 14                    & 242                         & 25                    & 19                        & 3                   & 204                  & 21            & 69                    & 11             & 15                  & 4            & 263                       & 26                  & 123                       & 16                  \\
\textbf{Multi-Object   Images}   & 326                         & 32                   & 345                         & 34                    & 209                         & 22                    & 244                       & 42                  & 236                  & 23            & 209                   & 33             & 166                 & 44           & 188                       & 19                  & 202                       & 27                  \\
\textbf{Single-Object   Images}  & 0                           & 0                    & 0                           & 0                     & 0                           & 0                     & 0                         & 0                   & 0                    & 0             & 24                    & 4              & 1                   & 1            & 0                         & 0                   & 0                         & 0                   \\
\textbf{Mislabled   Images}      & 5                           & 1                    & 18                          & 2                     & 28                          & 3                     & 27                        & 5                   & 64                   & 6             & 14                    & 2              & 3                   & 1            & 54                        & 5                   & 50                        & 7                   \\
\textbf{All Images}              & 506                         & 50                   & 497                         & 50                    & 479                         & 50                    & 290                       & 50                  & 504                  & 50            & 316                   & 50             & 188                 & 50           & 505                       & 50                  & 375                       & 50                  \\ \hline
\end{tabular}}
\vspace{-0.5cm}
\end{table*}


As from Fig.\ref{I1} (third column) the four stages introduced in the previous section can be enforced following the faceted classification methodology, being in fact mapped one-to-one to its four phases, i.e., \emph{Pre-Idea Stage}, \emph{Idea Plane}, \emph{Verbal Plane} and \emph{Notational Plane}. 
In fact, the Pre-Idea Stage is \emph{causally} concerned with the detection of objects as substance concepts, which are then (visually) classified in the Idea Plane, while the Verbal and the Notational Plane provide a standard mechanism for linguistic labelling and alinguistic rendering of the substance concept hierarchy, respectively. The key property of faceted classification is the fact that the work in all the phases is guided by a dedicated body of \emph{canons} ensuring classificatory finesse \cite{2017-CC}, namely guidelines and principles which must be followed in the annotation process, thus enabling the generation of high quality annotations with the further added value of increasing \emph{explainability} \cite{2018-XAI}. \textit{Canons constitute the core over which the methodology introduced in this paper is based.}

For lack of space we concentrate on the Idea Plane, this being the layer where the core part of SGP many-to-many mapping problem is concentrated, namely the problem of how to create one-to-one mappings between visual and linguistic properties. The reader can consult \cite{2021-CAOS} for an overall view of the canons of the other stages and  \cite{UKC-CICLING} to see how these canons have been followed (with local mistakes) for the generation of a multi-lingual WordNet-like lexical resource\footnote{This resource can be navigated at \url{http://ukc.datascientia.eu/}. This site will be extended to allow for the navigation of images.}. 
%
%
%
We concentrate on three sets of canons of prominent relevance for what concerns building the visual subsumption hierarchy, namely:
\begin{enumerate}
    \item \textbf{Canons for the selection and succession of concepts}, focusing on how to select visual features in any given point in the hierarchy.
    \item \textbf{Canons for horizontal concept expansion}, focusing on how to generate siblings.
    \item \textbf{Canons for vertical concept expansion}, focusing on how to expand the hierarchy into progressively higher levels of detail.
\end{enumerate} 
Let us consider these three sets of canons in some detail.

\vspace{0.1cm}\noindent \textbf{Selection and Succession of concepts.} This set of canons norm as to how a particular visual characteristic should be selected as the visual differentia and how they should be applied in succession at different levels of abstraction. We mention three  such canons. The first is the canon of \emph{relevance} which norms that the selected differentia should be relevant to the \emph{purpose} of the classification. E.g.,  sound producing mechanisms such as taut strings, keyboards etc. are appropriate visual differentia if the purpose is to classify musical instruments as per affordances. Secondly, the canon of \emph{ascertainability} enforces that a visual differentia \emph{``should be definite and ascertainable"} \cite{SRR-67}. E.g, a \emph{truss rod} being visually insignificant can't be used as a visual differentia differentiating a guitar from other musical instruments. Finally, the canon of \emph{relevant succession} norms that the succession of visual differentia should be relevant to the purpose of the classification. E.g., we take \emph{number of taut strings} as the first visual differentia to differentiate, for instance, between guitar and koto, with respect to which the former has six strings whereas the latter has thirteen. Further, the presence or absence of \emph{input jack} can be used  as the second visual differentia to differentiate between, for instance, electric guitar and acoustic guitar.

\vspace{0.1cm}\noindent \textbf{Horizontal concept expansion.} This set of canons prescribe as to how the sibling substance concepts at a specific level of abstraction should be modelled. We focus specifically on the canon of \emph{exhaustiveness} which prescribes that sibling substance concepts in an array \emph{``should be totally exhaustive of their respective common immediate universes"} \cite{SRR-67}. It further adds that a newly encountered object should either be classified into one of the existing visual categories represented as substance concepts or as a new substance concept altogether. This is crucial for image annotation where, for instance, all the known varieties of stringed instrument (such as \emph{guitar}, \emph{koto} etc.) should be made sibling concepts of the parent \emph{`stringed instrument'} with the possibility that a newly designed variety of string instrument can be assigned to any of the existing concepts or be classified as a new one based on the introduction of a new visual differentia. 

\vspace{0.1cm}\noindent \textbf{Vertical concept expansion.} The final set of canons provide guidance as to how taxonomically clean paths can be modelled in the visual subsumption hierarchy. One such canon is the canon of \emph{modulation}, which prescribes that a chain should be modelled such that it should comprise one concept \emph{``of each and every order that lies between the orders of the first link and the last link of the chain"} \cite{SRR-67}, in other words, ensuring that there shouldn't be gaps or missing links in visual classification hierarchies. A direct justification of the canon vis-à-vis recognition comes from the established fact \cite{ROSCH-1976} that there are basic categories which are probabilistically most optimal to be perceptually recognized and can never be missed out (for example, for musical instruments, the category \emph{guitar} cannot be superseded to directly jump from string instrument to acoustic guitar).

There are two important observations. Firstly, the fact that though we have a detailed set of canonical principles for ensuring the visual subsumption hierarchy to be ontologically thorough, the task becomes particularly challenging due to the tradeoff between the appropriate vertical and horizontal choice in uniquely classifying an object (see \cite{get-specific}). Secondly, 
the canons, in association with the four staged annotation, do provide the guidelines for enforcing a \emph{quality control} infrastructure which can be exploited to design a high quality ImageNet-like dynamically extensible visual hierarchy.
%
%
%
The key point, also factoring in other phases, is that the faceted classification process, while (of course) not  eliminating human subjectivity, does provide the guidelines for \emph{enforcing a one-to-one mapping} between visual and linguistic properties.


\section{Experiments}
\label{S6}

\begin{table*}[htp]
\centering
\caption{Group 1 and 2 Annotation Results.}
\label{tab:3}
\vspace{-0.2cm}
	\resizebox{1\linewidth}{!}{%
\begin{tabular}{|ccclccccccccc|clccccccccc|}
\hline
\multicolumn{1}{|c|}{\multirow{2}{*}{\textbf{Index}}}  & \multicolumn{1}{c|}{\multirow{2}{*}{\textbf{GT3}}} & \multicolumn{1}{c|}{\multirow{2}{*}{\textbf{GT4}}}  &  \multicolumn{10}{c|}{\textbf{GT1 (Annotation via Differentia)}}                                                                                                                                      & \multicolumn{10}{c|}{\textbf{GT2 (Annotation via Labels)}}       \\ \cline{4-23}
\multicolumn{1}{|c|}{}               & \multicolumn{1}{c|}{}                    & \multicolumn{1}{c|}{}         &    \multicolumn{1}{c|}{\textbf{Differentia}}                          & \textbf{U$_{1.1}$}  & \textbf{U$_{1.2}$}  & \textbf{U$_{1.3}$}  & \textbf{U$_{1.4}$}  & \textbf{U$_{1.5}$}  & \textbf{U$_{1.6}$}  & \textbf{U$_{1.7}$}  & \multicolumn{1}{c|}{\textbf{U$_{1.8}$} } & \textbf{S.D.} &        \multicolumn{1}{c|}{\textbf{Categories}}          & \textbf{U$_{2.1}$} & \textbf{U$_{2.2}$} & \textbf{U$_{2.3}$} & \textbf{U$_{2.4}$} & \textbf{U$_{2.5}$} & \textbf{U$_{2.6}$} & \textbf{U$_{2.7}$} & \multicolumn{1}{c|}{\textbf{U$_{2.8}$}} & \multicolumn{1}{c|}{\textbf{S.D.}} \\ \hline
\multicolumn{1}{|l|}{1}    & \multicolumn{1}{c|}{50}    & 41   &    \multicolumn{1}{|l|}{with Sound Mechanism}    & 33          & 12          & 27          & 25          & 28          & 36          & 12          & \multicolumn{1}{c|}{18}          & 9.0623            & \multicolumn{1}{l|}{Musical Instrument}  & 16          & 42          & 100         & 17          & 27          & 20          & 19          & \multicolumn{1}{c|}{26}          & \multicolumn{1}{c|}{28.1929}        \\
\multicolumn{1}{|l|}{1\_1}    & \multicolumn{1}{c|}{50}      & 123    &  \multicolumn{1}{|l|}{with Taut Strings}                   & 46          & 97          & 71          & 133         & 112         & 69          & 62          & \multicolumn{1}{c|}{79}          & 28.5354      & \multicolumn{1}{l|}{Stringed Instrument} & 162         & 78          & 0           & 115         & 144         & 106         & 161         & \multicolumn{1}{c|}{74}          & \multicolumn{1}{c|}{54.4085}        \\
\multicolumn{1}{|l|}{1\_1\_1}     & \multicolumn{1}{c|}{50}      & 34     &  \multicolumn{1}{|l|}{with 6 Strings}                       & 37          & 13           & 40          & 34          & 58          & 34          & 19           & \multicolumn{1}{c|}{31}          & 13.5831      & \multicolumn{1}{l|}{Guitar}              & 77          & 18          & 41          & 40          & 13          & 9           & 0           & \multicolumn{1}{c|}{8}           & \multicolumn{1}{c|}{25.4769}        \\
\multicolumn{1}{|l|}{1\_1\_1\_1}       & \multicolumn{1}{c|}{50}      & 40     &  \multicolumn{1}{|l|}{with No Input Jack}                     & 66          & 77        & 53          & 54          & 26          & 108        & 68          & \multicolumn{1}{c|}{50}          & 23.8253     & \multicolumn{1}{l|}{Acoustic Guitar}     & 13          & 81          & 66          & 43          & 70          & 70          & 82          & \multicolumn{1}{c|}{71}          & \multicolumn{1}{c|}{23.1393}        \\
\multicolumn{1}{|l|}{1\_1\_1\_2}     & \multicolumn{1}{c|}{50}     & 43   & \multicolumn{1}{|l|}{with Input Jack}                          & 54          & 61          & 67          & 43          & 28          & 12          & 82          & \multicolumn{1}{c|}{78}          & 24.2984    & \multicolumn{1}{l|}{Electric Guitar}     & 74          & 65          & 86          & 44          & 70          & 71          & 68          & \multicolumn{1}{c|}{74}          & \multicolumn{1}{c|}{11.8683}        \\
\multicolumn{1}{|l|}{1\_1\_2}     & \multicolumn{1}{c|}{50}      & 31   & \multicolumn{1}{|l|}{with 3 or 4 Strings} & 61          & 37          & 36          & 32          & 29          & 30          & 37          & \multicolumn{1}{c|}{30}          & 10.4335     & \multicolumn{1}{l|}{Dulcimer}            & 0           & 6           & 31          & 31          & 0           & 16          & 6           & \multicolumn{1}{c|}{46}          & \multicolumn{1}{c|}{17.1298}        \\
\multicolumn{1}{|l|}{1\_1\_3}     & \multicolumn{1}{c|}{50}      & 27    & \multicolumn{1}{|l|}{with 13 Strings}                         & 42          & 32          & 42          & 16          & 18          & 50          & 53          & \multicolumn{1}{c|}{45}          & 13.9668       & \multicolumn{1}{l|}{Koto}                & 0           & 47          & 47          & 39          & 0           & 41          & 0           & \multicolumn{1}{c|}{43}          & \multicolumn{1}{c|}{22.6239}        \\
\multicolumn{1}{|l|}{1\_2}     & \multicolumn{1}{c|}{50}     & 47     & \multicolumn{1}{|l|}{with Keyboard}                      & 49          & 46          & 41          & 49          & 43          & 47          & 50          & \multicolumn{1}{c|}{46}          & 3.1139     & \multicolumn{1}{l|}{Keyboard Instrument} & 41          & 49          & 0           & 44          & 40          & 47          & 52          & \multicolumn{1}{c|}{44}          & \multicolumn{1}{c|}{16.500}        \\
\multicolumn{1}{|l|}{1\_3}     & \multicolumn{1}{c|}{50}     & 47   & \multicolumn{1}{|l|}{with Embouchure}             & 62          & 60          & 60          & 63          & 50          & 48          & 60          & \multicolumn{1}{c|}{59}          & 5.5742        & \multicolumn{1}{l|}{Wind Instrument}     & 61          & 57          & 55          & 62          & 54          & 61          & 60          & \multicolumn{1}{c|}{59}          & \multicolumn{1}{c|}{2.9731}         \\
\multicolumn{1}{|l|}{}    & \multicolumn{1}{c|}{0}  & 17     & \multicolumn{1}{|l|}{Unrecognised}                         & 0           & 15          & 13          & 1           & 58          & 16          & 7           & \multicolumn{1}{c|}{14}          & 18.2757       & \multicolumn{1}{l|}{Unrecognised}        & 6           & 7           & 24          & 15          & 32          & 9           & 2           & \multicolumn{1}{c|}{5}           & \multicolumn{1}{c|}{10.4881}        \\ \hline
\multicolumn{1}{|l|}{all}   &   \multicolumn{1}{c|}{450}              & \multicolumn{1}{c|}{450}    & \multicolumn{9}{|l|}{Average S.D.}                                                                                                                                                              & 8.0470       & \multicolumn{9}{l|}{Average S.D.}                                                                                                       & \multicolumn{1}{c|}{13.2228}        \\ \hline
\end{tabular}
}
\vspace{-0.4cm}
\end{table*}

\begin{table}[t]
\centering
\caption{Group 1 and 2 Annotation Results - single object images.}
\label{tab:6}
\vspace{-0.2cm}
	\resizebox{1\linewidth}{!}{%
\begin{tabular}{|l|c|lccccccccc|}
\hline
\multicolumn{1}{|c|}{\multirow{2}{*}{\textbf{Index}}} & \multirow{2}{*}{\textbf{GT4}} & \multicolumn{10}{c|}{\textbf{GT1   (Only Single-Object images)}}                                                                                                                                             \\ \cline{3-12} 
\multicolumn{1}{|c|}{}                                &                               & \multicolumn{1}{c|}{\textbf{Categories}} & \textbf{U$_{1.1}$} & \textbf{U$_{1.2}$} & \textbf{U$_{1.3}$} & \textbf{U$_{1.4}$} & \textbf{U$_{1.5}$} & \textbf{U$_{1.6}$} & \textbf{U$_{1.7}$} & \multicolumn{1}{c|}{\textbf{U$_{1.8}$}} & \textbf{S.D.} \\ \hline
1                                                     & 17                            & \multicolumn{1}{c|}{with   Sound Mechanism}                  & 16            & 2             & 5             & 11            & 11            & 16            & 4             & \multicolumn{1}{c|}{4}             & 5.6045        \\
1\_1                                                  & 42                            & \multicolumn{1}{c|}{with Taut   Strings}                     & 21            & 41            & 32            & 43            & 41            & 33            & 28            & \multicolumn{1}{c|}{32}            & 7.4917        \\
1\_1\_1                                               & 21                            & \multicolumn{1}{c|}{with 6 Strings}                          & 21            & 8             & 23            & 19            & 26            & 29            & 11            & \multicolumn{1}{c|}{15}            & 7.2703        \\
1\_1\_1\_1                                            & 21                            & \multicolumn{1}{c|}{with No Input   Jack}                    & 31            & 34            & 29            & 28            & 18            & 41            & 33            & \multicolumn{1}{c|}{29}            & 6.5014        \\
1\_1\_1\_2                                            & 22                            & \multicolumn{1}{c|}{with Input Jack}                         & 24            & 32            & 31            & 21            & 20            & 0            & 39            & \multicolumn{1}{c|}{39}            & 12.7588        \\
1\_1\_2                                               & 13                            & \multicolumn{1}{c|}{with 3 or 4   Strings}                   & 24            & 13            & 13            & 14            & 12            & 10            & 15            & \multicolumn{1}{c|}{14}            & 4.1726        \\
1\_1\_3                                               & 12                            & \multicolumn{1}{c|}{with 13 Strings}                         & 12            & 7             & 10            & 8             & 9             & 9             & 13            & \multicolumn{1}{c|}{9}             & 1.9955        \\
1\_2                                                  & 33                            & \multicolumn{1}{c|}{with Keyboard}                           & 33            & 33            & 29            & 34            & 29            & 33            & 34            & \multicolumn{1}{c|}{33}            & 2.0529        \\
1\_3                                                  & 10                            & \multicolumn{1}{c|}{with Embouchure}                         & 20            & 20            & 21            & 23            & 14            & 12            & 19            & \multicolumn{1}{c|}{21}            & 3.7702        \\
                                                     & 11                            & \multicolumn{1}{c|}{Unrecognised}                            & 0             & 12            & 9             & 1             & 22            & 9             & 6             & \multicolumn{1}{c|}{6}             & 6.9166        \\ \hline
all                                                   & 202                           & \multicolumn{9}{l|}{Average S.D.}                                                                                                                                                            & 3.0080        \\ \hline
\end{tabular}}
\vspace{-0.5cm}
\end{table}

The experiments have been performed using the musical instruments ImageNet sub-hierarchy in Fig.\ref{I0}. As a first step, we use the proposed methodology to construct a ground truth dataset. 
Then, we describe an annotation experiment providing an evaluation of our proposed ground truth construction methodology. Finally, we compare the performance of various state-of-the-art ML algorithms, trained with various ground truths, including ours.

\subsection{Ground Truth Construction}
\label{exp.1}

Following the categorization in Section \ref{S2}, and applying the methodology in Section \ref{S3},  we have organized the images into the four categories  Good Images, Multi-Object Images, Single-Object Images, and Mislabelled Images. The final classification results are reported in Table \ref{tab:2} in the sub-column labeled ``Original", each column being associated with the corresponding category we are considering.

Some observations. The number of good images that meet our visual classification criteria, for each category, is quite variable and dependent on the category, always involving less than half of the total number of images, with multi-object images being always the biggest category and the single-objects images always the smallest. The ability of our methodology to identify mistakes is also highlighted by the fact that \cite{ICML-2020} discovered 270 mislabelled images out of around 10k images while we found a total of  264 out of around 3.66k images. As a case in point, though the image of a guitar-shaped wooden body without frets and taut strings can labelled as \emph{Guitar} (Fig.\ref{I1} (III)) following \cite{ICML-2020}, it is in our consideration a mislabelled image since it doesn't display any visual differentia such as `six taut strings'.



\subsection{Annotation Experiment}

The goal of this experiment was to evaluate how our methodology would perform when applied by non-expert annotators, as it is usually the case (given the size of the annotation tasks, the involvement of experts is unfeasible).  To this extent, we asked two groups of annotators to re-annotate the same subset of four hundred and fifty images from the ImageNet categories in Fig.\ref{I0} (fifty images per category). To minimize bias, the dataset was randomly generated, independently of the results in Table \ref{tab:2}. The quality of the resulting assignment, for each category is reported in the sub-column (labelled ``UE") in Table \ref{tab:2}. Furthermore, to minimize the personal bias, the  annotators were selected from geographically diverse backgrounds, e.g., Italy, Turkey, China, India etc., ensuring a good coverage of different cultures. Overall we produced four ground truths, each of four hundred and fifty images, organized as follows:
\begin{itemize}

\item GT1, as created by Group 1. The annotators were provided with a synthetic description of the methodology. The description was concentrated \textit{only} on Stage S2, namely on how to compute the differentia, this being the most important type of SGP and also the hardest to solve. 
They were asked to perform the annotation in two steps. First, they had to annotate images using the differentia, as extracted from the ImageNet gloss. In this step they were allowed to create a new category ``Unrecognized" for all the images they could not classify. Then, they were asked to label the nine categories with their ``most suitable" label, not necessarily from ImageNet.

\item GT2, as created by Group 2. This group was instructed to follow the ``usual" annotation process  where images would be tagged with the nine ImageNet category labels, based on their own experience. They were also allowed to create a tenth category ``Unrecognized".

\item GT3: the dataset, as categorized by ImageNet, where all the images are implicitly assumed to be good images. 

\item GT4: the dataset, as categorized by experts, with quality as from Table \ref{tab:2} sub-column ``Original". This was our ``reference" ground truth.
\end{itemize}

\vspace{0.1cm}\noindent
\textbf{Results and Analysis.}
A first set of results is reported in Table \ref{tab:3}. Here each line corresponds to a category, furthermore under GT1 we find its differentia, under GT2 we find the label naming it, namely the information passed to the two groups. The columns provide the number of images populating that Ground Truth. For GT1 and GT2 the numbers are provided per annotator, while the column ``S.D." reports the \textit{Sample Standard Deviation} across annotators.

The first key observation is the high variance with both GT1 and GT2. A rather negative result. However, if one looks closely at the last row, she will notice that in GT1 the average deviation is much lower than in GT2, being reduced by about 39\%. This by itself provides some evidence that our approach works. But things turn out definitely if one looks at Table \ref{tab:6} which reports the results only for single object images, thus not considering the noise introduced by multiple objects (which, according to our methodology are handled in Stage S1). Here it can be noticed that the average deviation decreases, with respect to GT2, of a factor of around 4.4, and with respect to GT1 of a factor of around 2.7, with an average value of 3.0080. And this result could be improved even more by eliminating the other two design flaws (see Section \ref{S2}). The interpretation of these results improves even more if we notice that U$_{1.6}$ is an outlier in that (s)he made a serious mistake in distinguishing the two categories ``1\_1\_1\_1" and ``1\_1\_1\_2", annotating a large number of images containing ``with Input Jack" as the ``with No Input Jack" category. If we fix this mistake, the S.D. of  these two categories in GT1 drops significantly to 15.3198 and 17.9523, with the Average S.D. going to 6.8699.

\begin{table}[t]
\caption{Category Labels suggested by Group 1 annotators.}
\label{tab:4}
\vspace{-0.2cm}
	\resizebox{1\linewidth}{!}{%
\begin{tabular}{|c|l|l|l|}
\hline
\textbf{User} & \textbf{Acoustic   Guitar}        & \textbf{Dulcimer}                                   & \textbf{Koto}                          \\ \hline
\textbf{U$_{1.1}$}          & Guitar                            & Appalachian   Dulcimer                              & Biwa                                   \\
\textbf{U$_{1.2}$}          & Guitar                            & IDK (I Don't Know)                                                 & IDK                                    \\
\textbf{U$_{1.3}$}          & Wooden   guitar                   & IDK                                                 & Zither                                 \\
\textbf{U$_{1.4}$}         & Wooden   guitar                   & 3 or 4 String Mucical Instrument             & 13 String Koto                 \\
\textbf{U$_{1.5}$}         & Hawaiian   Guitar                 & 3-4 Stringed Elliptical Instrument   & 13-String Instrument \\
\textbf{U$_{1.6}$}          & 6 Stringed Instrument no Jack     & Fretted Stringed Instrument                       & 13 Stringed Instrument         \\
\textbf{U$_{1.7}$}          & Classic   Guitar                  & Elliptical Stringed Instrument                    & Rectangular   Stringed Instrument      \\
\textbf{U$_{1.8}$}         & Non   Powered Guitars             & Short-Stringed Music Instruments                    & Japanese   Stringed Instrument        \\ \hline
\end{tabular}}
\vspace{-0.5cm}
\end{table}

A second observation is that in Group 2, but not in Group 1, some annotators were unable to annotate some images (see, e.g., the ``0"s in Table \ref{tab:3}). Two such examples are   ``Dulcimer" and ``Koto".\footnote{Dulcimers are popular in a region of the U.S. called Appalachia, while Koto is the national instrument of Japan.}
This observation is also confirmed in Table \ref{tab:4}, which reports some example labels provided during the second part of the Group 1 experiment. The observation here is that, despite properly labeling images via differentia, some annotators did not know the names of some categories, e.g., those of ``Dulcimer" and  ``Koto", or had in mind wrong or more generic labels, and in one case a more specific label. This provides further evidence that annotations via visual differentia improve the quality of annotations avoiding the problems of missing linguistic knowledge. 

%


All in all, these experiments\footnote{See the annotation results at: https://drive.google.com/drive/ u/1/folders/1Y3wDLh\_DYQAmIxu1wQElyCybtUqEHa3u} provide evidence of the pervasiveness of the SGP but also of how our stratified methodology allows to deal with it, one type of mistake at the time, following the process described in Section \ref{S4}. At the same it also shows, as also empirically pointed out in \cite{ILSVRC}, that it is crucial to enforce a \emph{quality control} which minimizes the probability of random mistakes by non-expert annotators. This is the next frontier for the scale up of high quality datasets.

\subsection{Machine Learning Experiment}

In this experiment we used four datasets DS1, DS2, DS3 and DS4. All four datasets contain the usual nine categories with a total of 1438 images, of which 1295 have been used as training sets and 143 as testing sets. The difference among the datasets was in the annotation process. DS1 was generated by general users according to the proposed methodology (similar to GT1), DS2 was generated by labelling categories (similar to GT2), DS3 consisted of the ImageNet labels (similar to GT3), while DS4 (similar to GT4) was generated by experts based on the proposed methodology.

All the experiments were implemented with PyTorch with identical settings. In the  training, we have performed the same data augmentation (random scaling and horizontal flipping) on each dataset and randomly sampled $224 \times 224$ crops from augmented images. All experiments were optimized using Adaptive Moment Estimation \cite{kingma2014adam}, with learning rate \cite{zulkifli2018understanding} initialized to $0.0002$, momentum initialized \cite{sutskever2013importance} to $0.9$, and weight decay \cite{loshchilov2018fixing} to $10^{-8}$. We used eight state-of-the-art ML methods to train the dataset collected by the three different annotation processes mentioned above, as listed in the first column of Table \ref{tab:5}. All models were trained from scratch with no pre-training.


\vspace{0.1cm}\noindent
\textbf{Results and Analysis.}
The experiment results are reported in Table ~\ref{tab:5}. The first column provides the list of the eight state-of-the-art ML methods which have been used.
Accuracies are reported in the second column.
All results consider the predicted label with the highest probability.
It can be noticed, that with the exception of AlexNet in DS1 vs. DS2, the accuracy improves substantially following the order DS3, DS2, DS1, DS4 thus confirming the validity of the methodology.
Note also that the results with DS1 are quite close to those computed with DS4. This confirms that it is fine to use non expert annotators, in particular if enforcing some quality control mechanism.

\begin{table}[t]
\centering
\caption{Classification results for the four datasets.}
\vspace{-0.2cm}
\label{tab:5}
	\resizebox{1\linewidth}{!}{%
\begin{tabular}{|l|cccc|}
\hline
\multicolumn{1}{|c|}{\multirow{2}{*}{\textbf{Method}}} & \multicolumn{4}{c|}{\textbf{Accuracy}}                    \\ \cline{2-5} 
\multicolumn{1}{|c|}{}                                 & \textbf{DS1} & \textbf{DS2} & \textbf{DS3} & \textbf{DS4} \\ \hline
AlexNet \cite{alexnet}          & 0.587                            & 0.594                            & 0.510                            & 0.608                                 \\
ZFNet \cite{zfnet}              & 0.664                            & 0.657                            & 0.608                            & 0.678                                  \\
VGG16 \cite{vgg}                & 0.748                            & 0.734                            & 0.699                            & 0.755                                  \\
GoogleNet \cite{googlenet}      & 0.818                            & 0.804                            & 0.727                            & 0.825                                  \\
ResNet 18 \cite{resnet}         & 0.727                            & 0.706                            & 0.538                            & 0.734                                  \\
DenseNet \cite{densenet}        & 0.769                            & 0.741                            & 0.692                            & 0.783                                  \\
Residual Attention Networks \cite{ran}      & 0.755                & 0.748                            & 0.706                            & 0.776                                  \\
SENets \cite{senet}             & 0.790                            & 0.783                            & 0.734                            & 0.804                                   \\ \hline
\end{tabular}
}
\vspace{-0.5cm}
\end{table}

\section{Related Work}
\label{S7}

As far as we know, this is the first time that a general full-fledged KR methodology is proposed, whose main goal is to produce high quality media datasets to be used for training and benchmarking CV algorithms. Still, in this work, we heavily rely and build on top of the ImageNet work~\cite{IMAGENET-2009}. With respect to this work the main innovations are as follows. First, the idea of using properties, and not only class names, for labeling images. Second, the idea of defining an annotation methodology, rather than just producing a data set, ready to be used to produce future high quality resources. A relevant issue is that the proposed methodology is fully incremental and can be used to improve, both in size and in quality the existing resources, ImageNet included. Third,  the exploitation of the faceted classification approach as a powerful technique for further improving the annotation quality. Finally, the fact that our approach allows for the generation of language aware annotations. A relevant issue here, which is a consequence of the work on the development of multi-lingual lexical resource which this work builds upon, is that the labels across language with the same meaning are all connected, still dealing with the well known untranslatability problems which exist when moving from one language to another, including the presence of lexical gaps \cite{UKC-CICLING}.

As also hinted in the introduction, the work proposed here constitutes also a solution to the SGP, restricted to how it appears in datasets. As shown in \cite{SNCS-2021} this is the first necessary step for development of algorithms which do not suffer from this problem. The work in \cite{SNCS-2021} is a first step in this direction. Our approach
to the solution of the SGP is also quite novel. Earlier work has focused on how to integrate feature-level information with semantic level information. Thus, some  have proposed using ontologies \cite{hare2006mind}, others have proposed to use high-level features \cite{ma2010bridging,elahi2017exploring}, others have proposed to ask users, also using active learning \cite{tang2011semantic}. More recently  \cite{pang2019} has proposed to handle the semantic gap in Deep Neural Networks when aggregating multi-level features. The common denominator in all this work is that it focuses on object labels, linguistically defined, rather than on the alignment between the visual properties of objects, as represented in media, and their linguistic description. 

Finally, a fair amount of work has also been done trying to model objects
in a way which is compliant to how humans think about objects.
Most of this work, motivated by (Cognitive) Robotic applications has concentrated on identifying the function of objects see, e.g., \cite{dimanzo1989understanding,stark1991achieving,bogoni1995interactive,woods1995learning,pechuk2005function,levesque2008cognitive}. The key difference is that this work has concentrated on how to enable a meaningful interaction and collaboration between humans and machines and not, as it is the case in this work, on how to enforce a process where there is coherence between how humans and machines describe and name objects.

\section{Conclusion}
\label{S8}

In this paper we have proposed a general KR methodology for producing high quality ground truth media datasets. The motivation for this work lies in the need of overcoming some of the limitations of current CV systems, partly induced by the low quality of annotation datasets. This is only a first step. We believe in fact that using KR models as the main mechanism for informing which examples should be fed in ML models is the key for the develpoment of explainable and high performance ML systems \cite{2019gini}.

\newpage

\section*{Acknowledgement}
The research conducted by Mayukh Bagchi and Xiaolei Diao is funded by the project \emph{“DELPhi - DiscovEring Life Patterns”} funded by the MIUR (PRIN) 2017.
The research conducted by Fausto Giunchiglia
is funded by the European Union Horizon 2020 FET Proactive project ``WeNet - the Internet of Us", grant agreement Number 823783. 

\bibliographystyle{kr}
\bibliography{kr}

\end{document}